\documentclass[letterpaper, 10 pt, conference]{ieeeconf}  % Comment this line out if you need a4paper

\IEEEoverridecommandlockouts                              % This command is only needed if 
\usepackage{amsmath}
\usepackage{amssymb}
\usepackage{graphicx}
\usepackage{booktabs}  % 漂亮的表格横线
\usepackage{wrapfig}
\usepackage{eso-pic}
\usepackage{xcolor}
\definecolor{red}{rgb}{0.75, 0.0, 0.0}

\title{\LARGE \bf
CoBrush: A Hierarchical Planning Framework for Human–Robot Co-Painting
}
\author{Dantong Qin$^{1}$ Yike Guo$^{2}$ Qinlin Liu$^{1}$ Alessandro Bozzon$^{1}$ and Pan Wang$^{1,*}$% <-this % stops a space
\thanks{$^{1}$Dantong Qin, Qinlin Liu, Alessandro Bozzon, and Pan Wang are with the Faculty of Industrial Design Engineering,
        Delft University of Technology, 2628 CD Delft, The Netherlands
        {\tt\small D.Q.Dantong@tudelft.nl, Qinlinliu619@gmail.com, A.Bozzon@tudelft.nl, P.Wang-2@tudelft.nl}}%
\thanks{$^{2}$Yike Guo is with the Department of Computer Science and Engineering, Electronic and Computer Engineering, Hong Kong University of Science and Technology, Kowloon Hong Kong
        {\tt\small yikeguo@ust.hk}}%
\thanks{$^{*}$Corresponding author to this work.}%
}

\begin{document}

\newcommand{\nb}[3]{
  \fcolorbox{black}{#2}{\bfseries\sffamily\scriptsize#1}
    {\sf\small$\blacktriangleright$\textit{\textcolor{brown}{#3}}$\blacktriangleleft$}
}

\newcommand\ale[1]{\nb{Ale}{red}{#1}}

\maketitle

\AddToShipoutPictureFG*{%
  \AtPageLowerLeft{%
    \put(0,\LenToUnit{8mm}){%
      \makebox[\paperwidth][c]{%
        \parbox[b]{0.9\paperwidth}{%
          \centering
          \normalfont\fontsize{8}{9.5}\selectfont
          Accepted at the 2026 IEEE/RSJ International Conference
          on Intelligent Robots and Systems (IROS 2026).
          \par
        }%
      }%
    }%
  }%
}

\thispagestyle{empty}
\pagestyle{empty}

%%%%%%%%%%%%%%%%%%%%%%%%%%%%%%%%%%%%%%%%%%%%%%%%%%%%%%%%%%%%%%%%%%%%%%%%%%%%%%%%
\begin{abstract}

Embodied co-painting requires a robot to repeatedly update a shared physical canvas while human intent evolves over interaction. Existing reference-driven painters or reactive assistants are typically optimized for single-shot rendering or sketch completion, limiting their ability to sustain coherent multi-round collaboration or to construct complex, content-rich scenes over time. We present CoBrush, a hierarchical framework that formulates multi-round co-painting as a coordinated semantic, spatial, and execution process. By separating high-level intent inference from spatial grounding and stroke-level control, the system supports progressive scene development on real acrylic canvases. We evaluate the framework through real human–robot painting sessions, stress tests, and user studies. Compared to single-turn baselines, our approach achieves stronger semantic alignment, more stable spatial progression, and higher perceived plausibility of robot actions. These results demonstrate that structured multi-stage reasoning improves the coherence and robustness of interactive painting and supports the progressive development of content-rich physical artworks.

\end{abstract}

\section{INTRODUCTION}

% This high-level planning bridges the gap between human creativity and machine execution by framing painting as a sequence of spatially and semantically grounded actions. 这句话可用
Embodied co-creation requires a robot to repeatedly update a shared physical artifact under evolving human input. In co-painting, this means deciding what to add, where to place it, and how to physically realize each update on a canvas over multiple rounds. Unlike predefined drawing tasks with fixed objectives, collaborative painting unfolds as an open-ended process in which goals and visual structures gradually emerge through interaction.

Existing systems only partially address this setting. Human–AI co-drawing methods have mainly been explored in digital sketch environments, where the focus lies on stroke completion \cite{davis2015drawing,davis2016empirically,oh2018lead} or refinement \cite{karimi2019relating,ibarrola2023collaborative,guljajeva2022dream}. While some support multi-turn interaction, they are often limited to sparse, line-based sketches and lack the mechanisms to manage long-term compositional coherence. Specifically, these systems prioritize immediate reactive updates over a structured understanding of how semantic elements should evolve and integrate within a visually rich scene.
Furthermore, their interaction remains confined to virtual canvases. Robotic painting systems \cite{gao2020making, luo2018robot, low2022drozbot, gao2024human,lindemeier2015hardware, scalera2019watercolour,schaldenbrand2023frida}, in contrast, excel at physical execution, but remain largely confined to non-interactive workflows. Predominantly reference-driven, these approaches rely on a target image (or sometimes with a text) provided at the outset, which the robot then executes in a single, automated pass. While prioritizing rendering fidelity, they also lack the adaptive planning required to process ongoing human guidance. Consequently, there is a lack of structured frameworks that enable an embodied agent to organize decisions to sustain coherent multi-round collaboration on a real canvas, which also constrains how compositionally complex artworks can emerge over time.

\begin{figure}[t]
    \centering
    \includegraphics[width=0.9\linewidth]{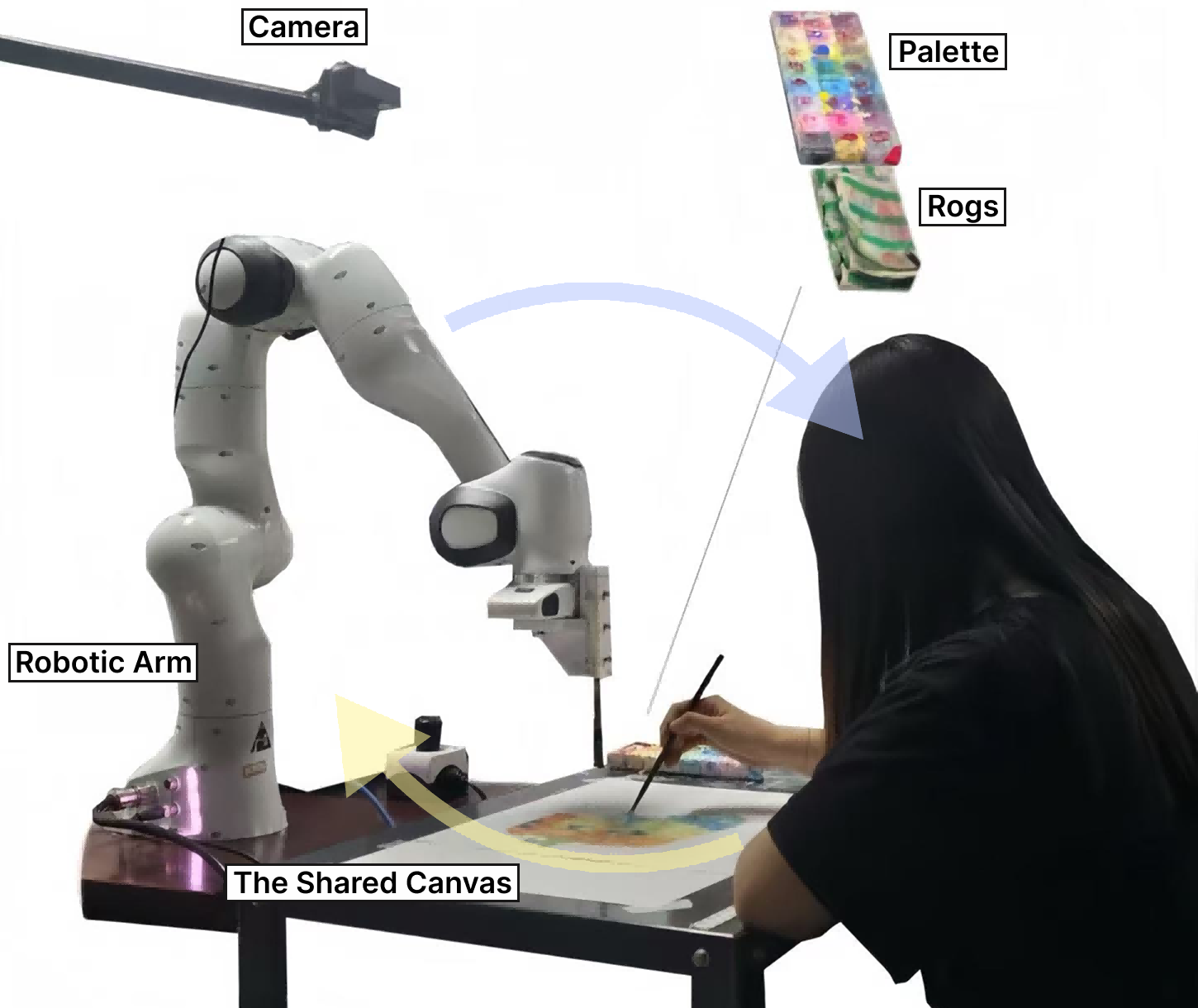}
    \caption{An overview of the human–robot co-painting setup, where a human and a robotic arm collaboratively paint on a shared canvas.}
    \label{fig:placeholder}
\end{figure}

In this work, we study multi-round, open-ended human–robot co-painting on a shared physical canvas, where the artwork progressively evolves through alternating interaction. We propose a hierarchical co-painting framework, \textbf{CoBrush}, that organizes the agent’s behavior around three complementary decision aspects, addressing what to depict, where it should appear, and how it is physically rendered through separate components within a unified embodied loop.
% that explicitly separates intent inference, spatial grounding, and physical execution within a unified embodied framework. 
To navigate the complexity of open-ended scenes, a vision–language model (VLM) first infers a high-level semantic target together with a coarse spatial prior from the evolving canvas, providing an initial hypothesis of both content and placement. This prior is subsequently refined through a coarse-to-fine localization process that leverages test-time weak supervision over visual features to correct VLM-induced spatial drift and prevent over-coverage. Finally, the refined region is translated into a sequence of executable brush strokes through a stroke-based painting policy. The robotic arm then physically realizes these updates on the acrylic canvas, bridging abstract high-level intent with concrete, physically realizable actions to complete the interaction round.

We evaluate CoBrush through real multi-round human–robot painting sessions, comparing it with representative single-shot approaches under the same interaction settings. The evaluation spans both process- and outcome-level perspectives, combining standard quantitative metrics with a stress test for robustness and a user study for collaborative quality. Across these complementary measures, results consistently indicate improved semantic alignment, more accurate spatial localization, and higher perceived plausibility of robot actions, suggesting that iterative perception and planning over interaction rounds supports more coherent open-ended collaboration. 

\section{Related Work}

\subsection{Human-AI collaborative drawing system}

Human–AI co-creation has been widely studied in digital sketching contexts. Early dialog-based systems enabled multi-turn stroke-level interaction with limited semantic understanding \cite{davis2015drawing,davis2016empirically}. Sequential models such as Sketch-RNN \cite{oh2018lead,ha2017neural} formalized stroke prediction for object-level sketch completion. More recent approaches leverage multimodal alignment and diffusion-based editing. DreamPainter and ReFramer incorporate CLIP-based \cite{radford2021learning,frans2022clipdraw} vision–language guidance to support voice-conditioned or iterative sketch refinement \cite{guljajeva2022dream,ibarrola2023collaborative}, while CoFRIDA employs controllable diffusion for collaborative sketch updating \cite{schaldenbrand2024cofrida}. While being interactive, these systems are largely reactive: the AI primarily serves to complete, refine, or embellish user input, rather than actively reasoning about the evolving composition or contributing at a structural level. This necessitates advanced planning in human–robot co-painting systems.

\subsection{Vision-language models}

Vision–language models (VLMs) have recently been integrated into embodied agent systems, serving as high-level perception and reasoning modules for robotic manipulation \cite{pan2025omnimanip}, embodied navigation \cite{ziliotto2025tango}, creative design \cite{feng2023layoutgpt,lin2025elements,xue2025comfybench}, and controllable image generation \cite{tang2025ata}. In collaborative drawing, SketchAgent \cite{vinker2025sketchagent} directly uses a frozen VLM to generate grid-based stroke coordinates from language prompts, tightly coupling semantic reasoning with low-level stroke generation in a digital sketch environment. Following this trend, our system employs a VLM for high-level semantic planning, while delegating localization and stroke execution are handled by dedicated modules.

% Recent advances in multimodal foundation models and instruction-tuned agents have opened up new possibilities for human-AI collaboration.  
\begin{figure*}[t]
    \centering
    \includegraphics[width=1\linewidth]{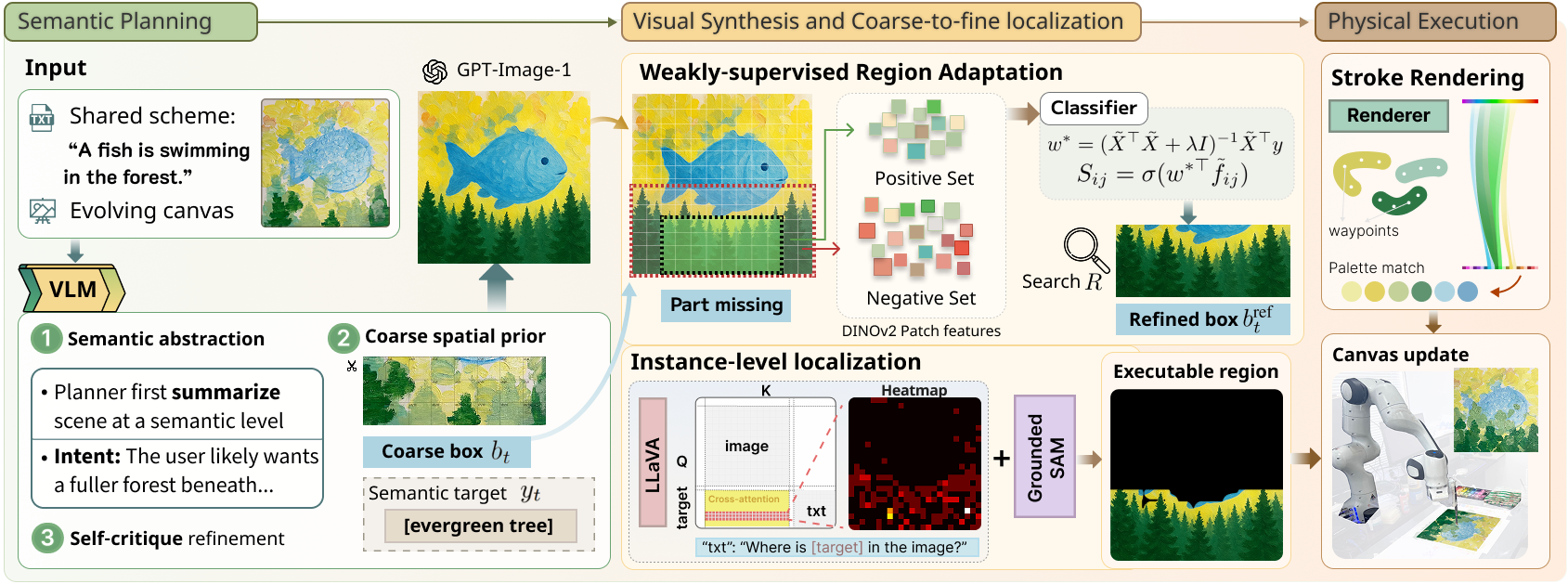}
    \caption{Overview of the CoBrush. The system iteratively connects semantic planning, visual synthesis with coarse-to-fine localization, and physical execution to transform a shared theme and evolving canvas into executable painting actions on a real canvas.}
    \label{fig:agent}
    % \vspace{-2mm}
\end{figure*}

\subsection{Robotic painting}

Robotic painting has been widely studied in autonomous portrait sketching and non-photorealistic rendering \cite{gao2020making,luo2018robot,low2022drozbot,gao2024human,jin2023semi}, as well as 3D surface drawing \cite{liu2021robust, song2023ssk} and calligraphy \cite{zhang2019intelligent}. Most of these systems focus on reference-driven or predefined objectives, typically operating in single-shot settings. Beyond monochrome line drawing, colored robotic painting has been explored in \cite{lindemeier2015hardware,scalera2019watercolour}. More recently, FRIDA \cite{schaldenbrand2023frida} leverages multimodal generative models for text- and image-conditioned robotic painting, while CoFRIDA \cite{schaldenbrand2024cofrida} applies diffusion-based editing for iterative refinement on a user-provided canvas. However, these systems lack structured planning for evolving guidance, which restricts their interactive flexibility. We instead treat each round as a decision-making process to ensure reliable intent-to-stroke mapping, which effectively enables continuous multi-round collaboration.

\section{Method}
We present CoBrush, a hierarchical embodied framework for multi-round human–robot co-painting (Fig.~\ref{fig:agent}). The method is organized into three interconnected modules: a semantic planner for high-level intent inference, a grounding-and-synthesis module for spatial localization, and a stroke-based executor for physical rendering. We then describe how these components are integrated within an iterative human–robot collaboration loop.

\subsection{CoBrush: Semantic planning}

To determine what to paint next and where to paint it, we employ a semantic–spatial planner implemented using a frozen vision–language model (VLM), such as GPT-4o. We structure the VLM’s behavior through instruction prompts that transform the evolving canvas and the theme caption (both from human) into an actionable prediction. At each interaction round $t$, the planner produces a structured output $(y_t, b_t)$, where $y_t$ is a high-level semantic target describing what content to add, and $b_t$ is a coarse spatial prior indicating where this content should be placed on the canvas.

\subsubsection{Semantic abstraction} the planner first summarize the scene at a semantic level. Specifically, the VLM generates two complementary descriptions:
(i) a visual summary that captures the current layout and dominant elements on the canvas, and
(ii) an intent-oriented summary that predicts a plausible next development under the given theme.
Based on these abstractions, the VLM synthesizes a concise semantic target $y_t$ (e.g., “eyes and nose”), together with a brief justification explaining how this addition advances the artwork.

\subsubsection{Coarse spatial prior} Direct coordinate regression is unreliable for VLMs \cite{pan2025omnimanip,vinker2025sketchagent}, particularly in open-ended scenes where precise pixel locations are ambiguous. we therefore cast region prediction as a discrete selection task. The canvas is overlapped with a $10 \times 10$ grid, and the VLM outputs a bounding region $b_t = [idx, w, h]$, where $idx$ specifies the top-left grid cell and $(w, h)$ denote the region extent in grid units. The discretization reduces spatial estimation to low-cardinality classification, improving stability across different canvas resolutions.

\subsubsection{Self-critique refinement} To further improve robustness, we incorporate a lightweight self-critique step. The same VLM re-evaluates whether the predicted region aligned with intended composition by assigning a binary status (answering ``Yes'' or ``No'') based on the visible grid overlay, and provides a short rationale. When the prediction is rejected, the VLM is prompted to revise it. This loop typically runs for one to two iterations and helps reduce common spatial errors like misalignment.

\subsection{CoBrush: Visual synthesis and Coarse-to-fine localization}
Given the semantic target $y_t$ and coarse prior $b_t$, this module bridges high-level intent and low-level execution by synthesizing a visual reference and refining its spatial extent into a precise painting region for rendering.

\subsubsection{Visual synthesis}
The semantic target $y_t$ (e.g., ``eyes and nose'') is often too abstract to guide spatial grounding and stroke rendering directly. We therefore instantiate a visual reference that illustrates how the missing content is expected to appear on the current canvas. Using the visual summary and the target phrase $y_t$, we prompt GPT-Image-1 to generate an exemplar image $I_t^{\text{ref}}$ that aligns with both the intended content and the local painting style. 
% It serves as a reference for the subsequent localization and rendering stages.

\subsubsection{Weakly-supervised region adaptation}
The coarse box $b_t$ predicted by planner provides a useful prior but often suffers from spatial drift or over-/under-coverage. To refine it into a more accurate painting region, we treat $b_t$ as weak supervision and exploit local visual features at test time. Specifically, we use the region inside $b_t$ as a noisy positive set, while a thin ring around the box serves as a local negative set.
This construction allows us to learn a lightweight classifier that discriminates which part of the coarse region actually corresponds to the target. Given $I_t^{\text{ref}}$, we extract DINOv2 \cite{oquab2023dinov2} patch features $F \in \mathbb{R}^{H \times W \times D}$, where each patch feature is denoted by
$f_{ij} \in \mathbb{R}^{D}$ at grid location $(i,j)$. We sample features from the positive set $\mathcal{P}$ (inside $b_t$) and the negative set $\mathcal{N}$ (the surrounding ring).
Let $X$ denote the concatenated features and $y \in \{0,1\}$ their weak labels. We fit a Ridge regression probe with a bias by augmenting each feature with a constant $1$.
Let $\tilde X = [X\ \mathbf{1}]$. The closed-form solution is:
\[
w^\ast = (\tilde X^\top \tilde X + \lambda_{\text{ridge}} I)^{-1}\tilde X^\top y,
\]
where $I$ is the identity matrix. The resulting classifier assigns each patch a relevance score:
\[
S_{ij} = \sigma(w^{\ast\top} \tilde f_{ij}),
\]
where $\sigma(\cdot)$ is the sigmoid function. We then search for the rectangle $R$ that maximizes the accumulated relevance while penalizing excessive area:
\[
b_t^{\text{ref}} = \arg\max_R \left(
\sum_{(i,j)\in R} S_{ij} - \lambda_{\text{area}} |R|
\right).
\]
We denote the refined bounding box as $b_t^{\text{ref}}$, which replaces the coarse prior $b_t$ and serves as the final rectangular region for instance-level localization. This refinement is particularly useful when the coarse localization suffers from typical VLM errors, such as covering only part of a semantic target (e.g., half of a cat’s head) or including large amounts of irrelevant background. As DINOv2 features are highly linearly separable, the classifier is both fast and reliable, enabling real-time refinement in an interactive painting scenario.

\subsubsection{Instance-level localization}

While the refined box $b_t^{\text{ref}}$ identifies the accurate region, stroke planning requires a finer instance mask. We therefore combine complementary cues from a multimodal LLM and a segmentation model to obtain it. Given the target phrase $y_t$, we use LLaVA-1.6 (Mistral-7B) \cite{liu2023visual} to compute text–image cross attention over the canvas, producing a semantic heatmap that highlights areas most relevant to $y_t$. Grounded-SAM \cite{ren2024grounded} also generates class-agnostic instance masks with precise geometric boundaries, complementing the heatmap’s irregular contours. Their union yields a target-consistent and geometrically complete painting region for rendering.

\subsection{CoBrush: Physical execution}
Given the refined painting region and the target-conditioned reference image, we translates the desired update into a sequence of physically executable strokes and colors for robot painting.
\subsubsection{Stroke rendering}
We adopt StrokeDiff \cite{qin2026data} as a stroke renderer to converts pixel-based content into executable primitives.
For each selected region, the renderer decomposes the reference image into a collection of cubic Bézier strokes parameterized by control points and RGB color. Each Bézier curve is uniformly sampled into 2D waypoints on the canvas plane and lifted to 3D end-effector poses through the calibrated camera–canvas transformation, producing trajectories that can be directly tracked by the robot using standard Cartesian velocity control.
% We adopt a stroke renderer established on \cite{liu2021paint} to converts pixel-based content into executable primitives.
% For each selected region, the renderer decomposes the reference image into a collection of cubic Bézier strokes parameterized by control points and RGB color. Each Bézier curve is uniformly sampled into 2D waypoints on the canvas plane and lifted to 3D end-effector poses through the calibrated camera–canvas transformation, producing trajectories that can be directly tracked by the robot using standard Cartesian velocity control.

\subsubsection{Palette-aware robot execution} 
While StrokeDiff operates in continuous RGB space, the physical setup is constrained to a discrete acrylic-paint palette. We therefore quantize stroke colors in HSV space into a small set of artist-friendly bins {\small\texttt{[red, orange, yellow, green, blue, purple, pink]}} and three value levels {\small\texttt{[light, medium, dark]}} to reduce color contamination, resulting in 21 palette entries matched to the physical mixing tray. Each stroke is assigned to its nearest palette bin, producing a compact, robot-readable color label. Each waypoint carries its palette label; when a color change is required or paint is depleted, the arm moves to predefined dipping and cleaning poses before resuming painting. This palette-aware policy allows the model to approximate continuous color choices using a small and reusable set of physical paints.

\subsection{Human–Robot Collaboration}

At the beginning of a co-painting session, the human provides a shared theme caption that specifies the overall subject and style of the artwork and remains fixed throughout the session. The collaboration follows a turn-based closed loop. After each human update, an overhead camera captures the current canvas as CoBrush’s input. Conditioned on the shared theme and the observed canvas, CoBrush generates a new painting update and executes it on the canvas. Control then returns to the human, who can react to, modify, or extend the result. This alternation continues over multiple rounds, allowing the artwork to gradually emerge through iterative dialogue.

\section{Experiment }

\begin{figure*}
    \centering
    \includegraphics[width=1\linewidth]{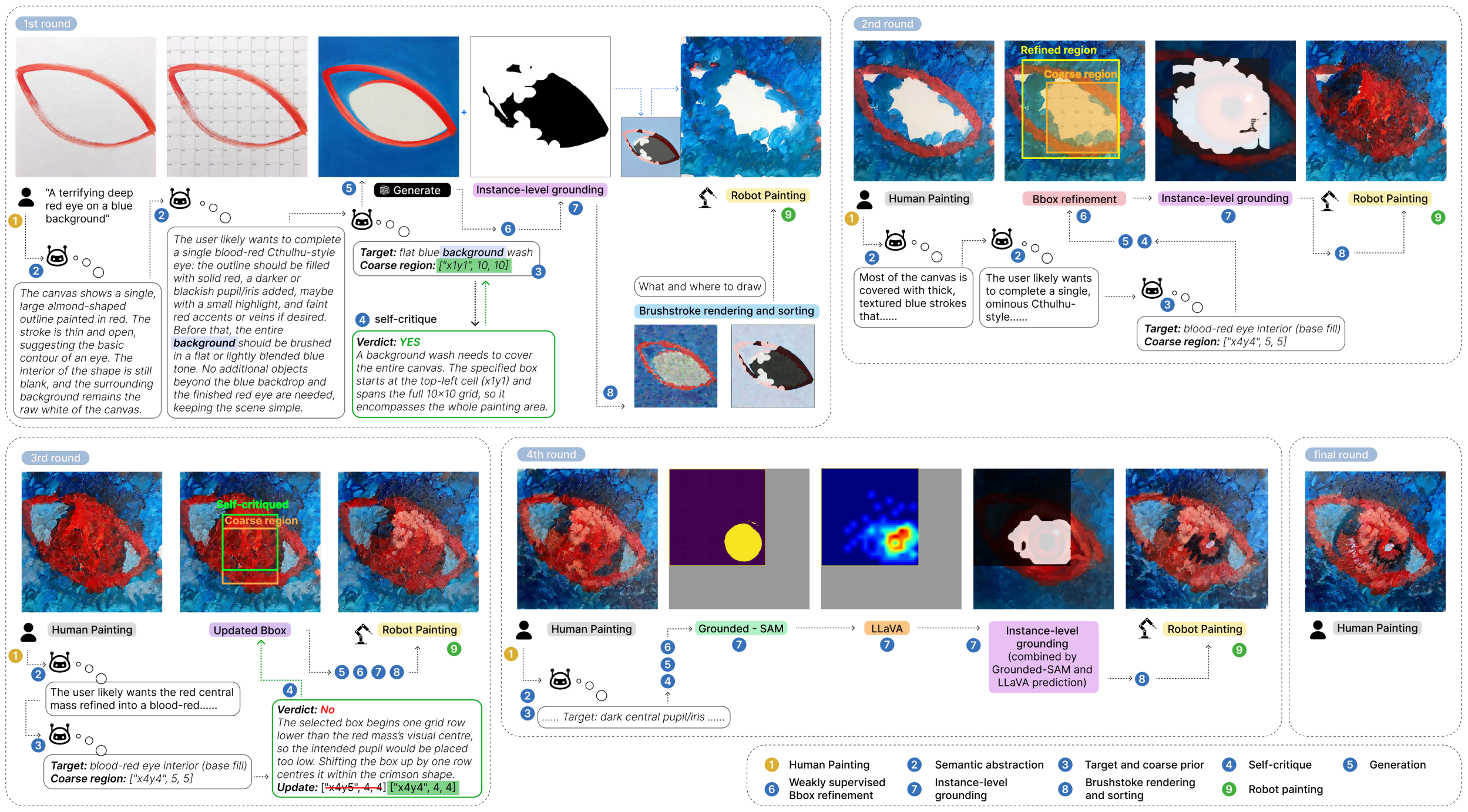}
    \caption{Example of a multi-round co-painting session. The figure illustrates alternating human and system contributions across several rounds, along with the system’s semantic planning, region refinement, grounding, and stroke execution steps.}
    \label{fig:result_detail}
\end{figure*}

\begin{table*}
\centering
\caption{Comparison across all methods.
% Metrics include PSA (Intent), OSA (Outcome), and SEC (Confidence) for semantic evaluation, alongside P-IoU and Posterior Rationality scores (target/bbox) for spatial and interactive plausibility.
Higher values indicate better alignment or plausibility except where noted. }
% \ale{Our system has a name. Use the name}}
\label{tab:quantitative}
\begin{tabular}{lccccccccc}
\toprule
& \multicolumn{1}{c}{PSA} 
& \multicolumn{2}{c}{OSA} 
& \multicolumn{2}{c}{SEC} 
& \multicolumn{1}{c}{P-IoU} 
& \multicolumn{2}{c}{Post. anchor} \\
\cmidrule(r){2-2} \cmidrule(r){3-4} \cmidrule(r){5-6} \cmidrule(r){7-7} \cmidrule(r){8-9}
Method 
& CLIP
& GPT-4.1 & LLaVA-OV 
& Yes $\uparrow$ & No $\downarrow$
& IoU
& target & bbox \\
\midrule
CLIPDraw 
& --    & 53.25          & 52.36          & 0.521 & 0.478 & --      & --   & 5.9   \\
Frida 
& --    & 50.42          & 34.00          & 0.487 & 0.512 & --     & --   &  4.1  \\
CoBrush (w/o multi) 
& --    & 47.30          & 25.14          & 0.388 & 0.611 & --      & --   &  6.0  \\
Cobrush 
& 0.675 & \textbf{70.75} & \textbf{63.75} & \textbf{0.668} & \textbf{0.335} 
& 0.376            & 7.5  & \textbf{8.1} \\
\bottomrule
\end{tabular}
\end{table*}

\subsection{Implementing details}
% \noindent\textbf{Implementing details.} 
We deploy the system on a 7-DoF Franka robotic arm equipped with a fixed brush holder, and use a simple sponge pad for brush cleaning between color switches. To reduce redundant motions, strokes of the same color with over 80\% spatial overlap are pruned before execution. Although it introduces some geometric loss, it reduces execution time by more than 5$\times$ in practice.

\subsection{Participants}
% \noindent\textbf{Participants. }
We recruited 14 participants (9 female, 5 male), aged 20–46 ($M = 28.64$, $SD = 7.83$), through online channels and local community mailing lists. The study was approved by our institutional review board (IRB). Participants came from diverse professional backgrounds, including semiconductor engineering, finance, law, aerospace, photography, art education, and art history, and were selected based on their interest in human–robot co-painting. Their painting experience ranged from 0 to 12 years ($M = 4.9$ years), with two participants reporting no prior painting experience. Two participants had previous experience using a robotic arm, while two were entirely unfamiliar with robotics. Each participant completed the study individually, with a typical session lasting 4-5 hours. Among the 14 participants, two were unable to finish their paintings due to scheduling constraints, and one session was excluded due to a system failure, resulting in 11 completed artworks used for evaluation.

\subsection{Compared Methods}
% \noindent\textbf{Compared Methods. }
We compare our system with two representative open-source text-guided drawing and painting methods, as well as an internal ablation. These methods are selected because they are among the closest publicly available systems that support robotic or stroke-based image synthesis from language. However, none of them is designed for multi-round human–robot collaboration. Therefore, all baselines are evaluated in a single-shot setting. 1) \textbf{CLIPDraw} \cite{frans2022clipdraw} synthesizes line-based drawings from natural language prompts through CLIP-guided stroke optimization and has been adopted in a robotic drawing system \cite{guljajeva2022dream}. To ensure a fair comparison, we initialize the optimization from the participant’s first-round canvas, allowing CLIPDraw to generate an update based on the same visual context as our system. 2) \textbf{FRIDA} \cite{schaldenbrand2023frida}. In our experiments, we use FRIDA in its text-conditioned mode only. Similar to CLIPDraw, FRIDA is applied to the participant’s first-round canvas to produce a single painting update for comparison. 3) \textbf{Ours (w/o. multi-turn)}.
To separate the effect of multi-round reasoning, we include an ablated variant of our system that removes semantic planning and region prediction. This variant executed a single update based solely on the shared scheme and initial canvas, without incorporating any iterative human feedback.

\subsection{Metrics}
% \noindent\textbf{Metrics. }
We evaluate the system using a hybrid of automatic and user-centric metrics.
% measure semantic and spatial accuracy, with task-specific metrics \ale{Which ones are task-specific? DId we create them>} capture whether the system’s actions are perceived as reasonable during multi-round collaboration.

\textit{Semantic Intent Interpretation. }1) {\textbf{Process-level semantic Alignment (PSA).} This metric measures whether the system anticipates the user’s intent during interaction. For each round, we compare the planner’s predicted next-step target with the user-provided target using CLIP similarity, reflecting how well the robot aligns with the user’s intended content before execution. 2) \textbf{Outcome-level semantic Alignment (OSA).} As PSA only can be applied to CoBrush, this metric evaluates whether the intended semantic target is actually realized in the final artwork. For all methods, we assess whether the user-specified target appears within the user-provided bounding box in the completed image. Following the LLM-as-a-Judge in T2I-CompBench++ \cite{huang2025t2i}, we prompt GPT-4.1 and LLaVA-OneVision with a chain-of-thought instruction (e.g., ``Rate from 0 to 100 how likely there is a [target] inside the bounding box.''). The resulting likelihood score reflects semantic alignment in the completed artwork. 3) \textbf{Semantic Existence Confidence (SEC). }This metric provides a supplementary check of semantic correctness. We ask LLaVA-OneVision a constrained yes/no question (e.g., “Is there a [target] inside the image?”) and compute the normalized logit difference between the “yes” and “no” tokens. This value serves as a calibration-oriented confidence score, complementing the continuous 0–100 ratings. 4) \textbf{Posterior rationality (target-anchored). }As painting is non-deterministic and multiple next choices may be reasonable, we also evaluate whether the system’s predicted target is perceived as plausible by the participant. After each round, users rate on a 0–10 scale, how reasonable the predicted target is as a next step given the current canvas and theme.

\textit{Spatial Localization Accuracy. }1) \textbf{Procedural IoU (P-IoU).} This metric evaluates whether the system places its actions in the intended region during interaction. For each round of our system, we compute the intersection-over-union (IoU) between the predicted next-step region and the bounding box specified by the participant, measuring how accurately the robot follows the user’s spatial guidance. 2) \textbf{Posterior rationality (bbox-anchored). }This metric captures spatial plausibility beyond strict geometric overlap. Since multiple placements may be acceptable in creative painting, participants rate, on a 0–10 scale, how reasonable the robot’s generated content is within the region they specified, given the current canvas and the shared theme.

\begin{figure*}
    \centering
    \includegraphics[width=1\linewidth]{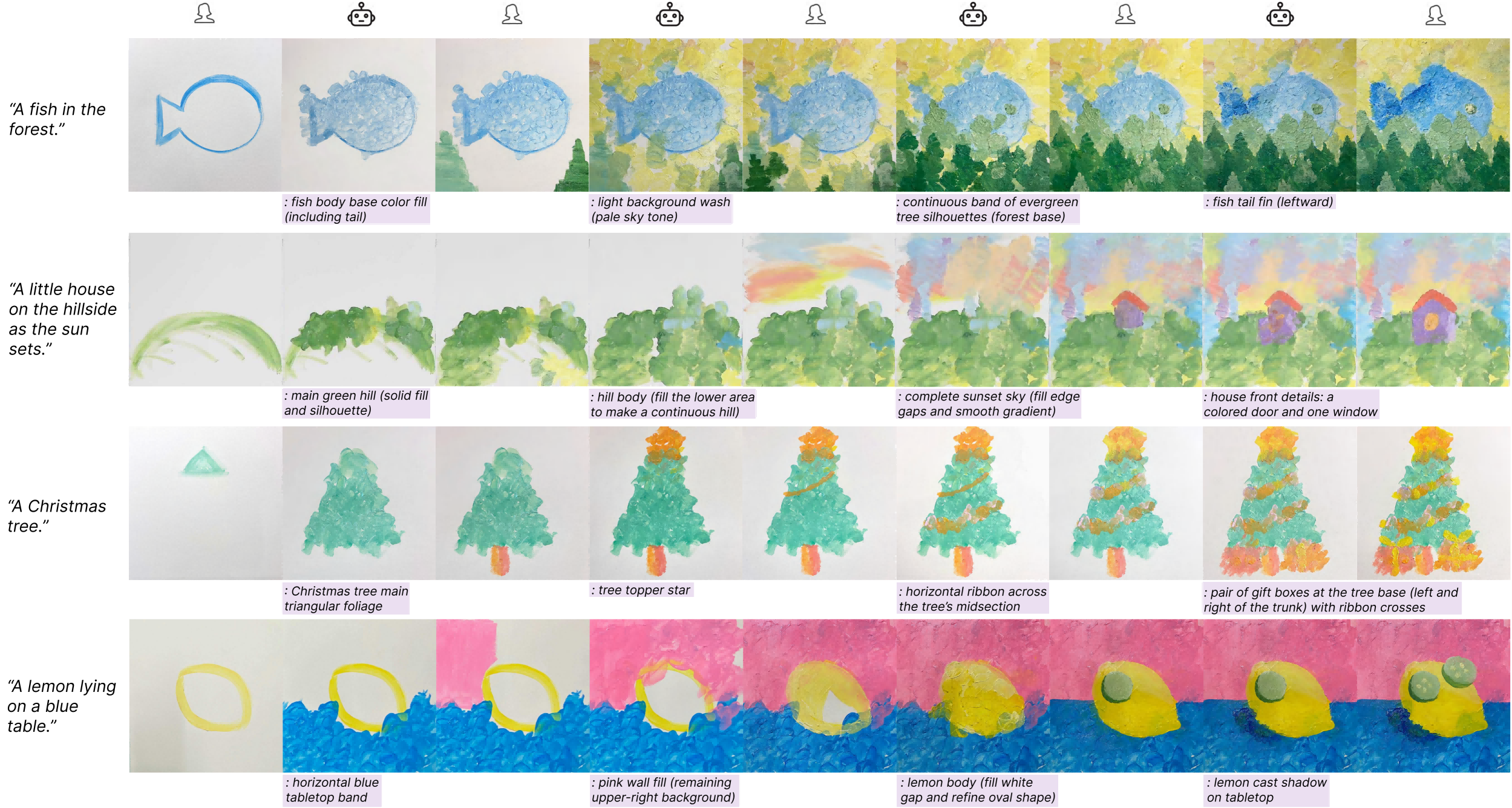}
    \caption{Multi-turn co-painting examples across different user themes. Human and agent contributions alternate from left to right. Text under each agent update indicates the system’s predicted next-step target.}
    \label{fig:main_result}
        \vspace{-2mm}
\end{figure*}

\begin{figure}
    \centering
    \includegraphics[width=1\linewidth]{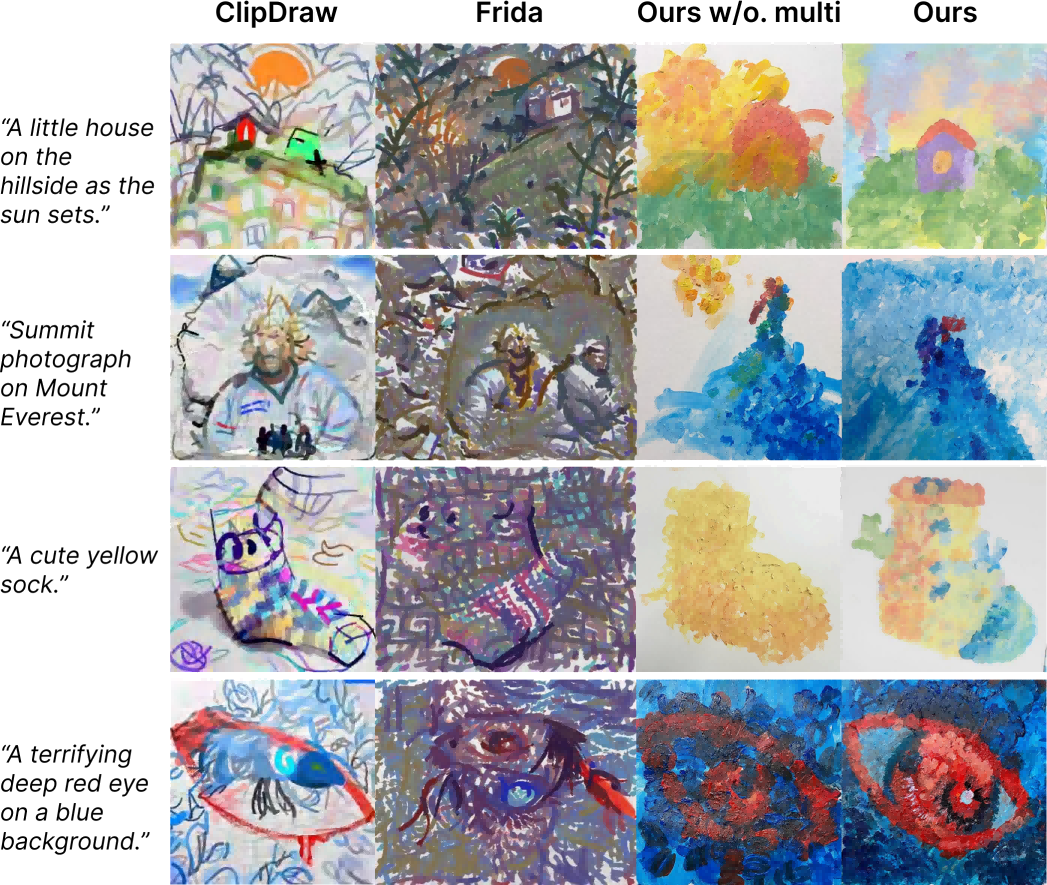}
    \caption{Comparison of final paintings using four methods: CLIPDraw \cite{frans2022clipdraw}, FRIDA \cite{schaldenbrand2023frida}, our single-turn ablation, and our full multi-turn system. Each column shows the output generated from the same initial first-round canvas.}
    \label{fig:compare}
\end{figure}

\section{Results}

\subsection{Quantitative Results}
% \noindent\textbf{Quantitative results. }
CoBrush consistently outperforms baselines across all semantic and spatial dimensions. In terms of semantic interpretation, our system achieves a PSA of 0.675, and the best OSA among all methods, indicating that its planning aligns closely with user intent both as a predictive strategy and in the final artistic realization. The highest SEC values further confirm that intended subjects are more reliably integrated into the composition compared to single-turn approaches. Regarding spatial reasoning, CoBrush is the only method capable of dynamic localization, achieving a mean P-IoU of $0.376$, demonstrating reliable placement of future updates. Furthermore, the significantly higher Posterior Rationality ratings (for both target and bounding box) suggest that the agent’s decisions are perceived as more collaborative and plausible by human partners. Overall, these results show that incorporating a structured, multi-turn reasoning loop substantially enhances the robot's capacity to interpret and execute complex, evolving creative intents.

\subsection{Qualitative Results}
% \noindent\textbf{Qualitative results. }
Figure \ref{fig:result_detail} presents a full interaction trajectory, illustrating how the system interprets the evolving canvas, proposes semantically plausible next steps, refines spatial grounding, and executes targeted updates. The progression highlights the role of semantic abstraction and region refinement in maintaining coherent visual development across rounds. Figure \ref{fig:main_result} shows additional outputs produced under a variety of user themes. Across these scenarios, the next-step target under each update shifts with the canvas state, generating updates that integrate smoothly with the existing canvas. Figure \ref{fig:compare} provides comparisons with the baseline methods. While each baseline reflects its own rendering strategy and aesthetic characteristics, the resulting images demonstrate differences in how systems respond to the partially completed first-round canvas. Our full model produces updates confined to the targeted region that preserve earlier strokes, whereas the ablated single-turn variant reflects only the initial state without incorporating intermediate feedback. These comparisons highlight the distinct behaviors of single-turn versus iterative systems in interactive painting contexts.

\subsection{Region Refinement Effectiveness}
% \noindent\textbf{Region Refinement Effectiveness. }
To demostrate the effectiveness of our weakly-supervised rectangular region refinement, we compare the planner’s coarse region with the refined region produced by our DINOv2-based scoring method. As shown in Fig. \ref{fig:score_map}, when the planner’s coarse region does not fully capture the intended area, the refinement module adjusts the box using weak supervision. Features inside the coarse box are treated as positives and those in a surrounding ring as negatives, producing a score map that reflects visual similarity under this local discrimination objective. The rectangular search then selects a region that better follows the high-score areas. In the illustrated examples, the refined boxes expand or shift when the coarse box is insufficient, resulting in regions that more completely cover the visually coherent part associated with the indicated target concept.

\begin{figure}
    \centering
    \includegraphics[width=1\linewidth]{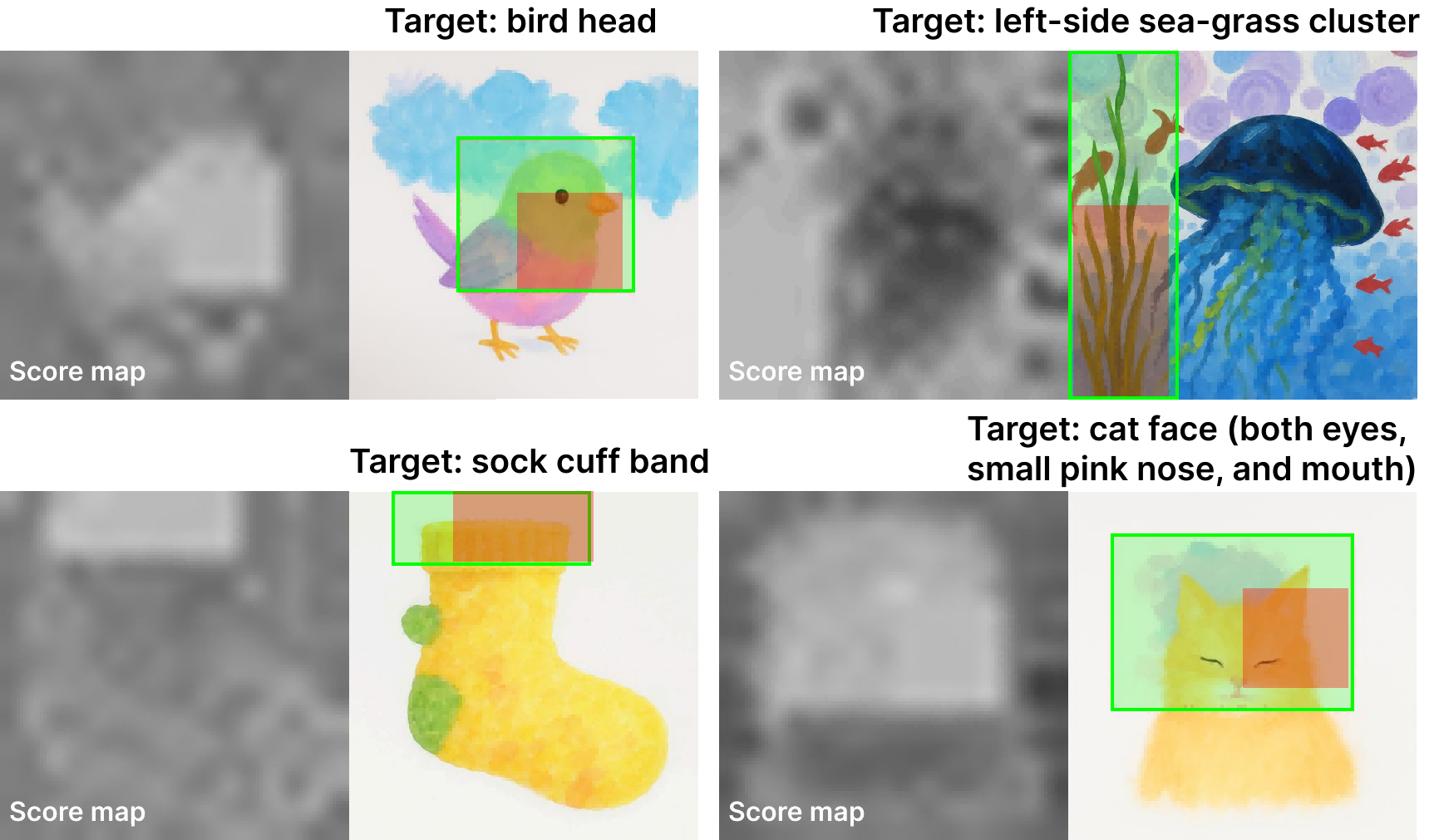}
    \caption{Effect of weakly supervised region refinement. Red boxes show the planner’s coarse spatial predictions $b_t$; score maps depict patch-level relevance from the DINOv2-based classifier; green boxes indicate the refined regions $b_t^{ref}$ selected by our search procedure.}
    \label{fig:score_map}
    \vspace{-1mm}
\end{figure}

\subsection{Goal Drift Stress Test}

To test the agent's limits beyond the "cooperative" settings of the main study, we introduce a goal drift stress test. This evaluates robustness when user input becomes ambiguous, creative, or intentionally inconsistent with the global theme. We selected 18 intermediate canvas states (sampled from the 11 sessions in the main study) and presented them to six new participants. For each selected state, we reproduced three identical canvas copies to independently test one step of interaction at three stress levels: participants were asked to 1) \textit{A-Mild:} add minor off-theme decorations; 2) \textit{B-Moderate:} add dominant but theme-inconsistent content; 3) \textit{C-Severe.} add strongly conflicting or destructive content that are difficult to reconcile with both the canvas and the theme.

We evaluate the system’s adaptability to challenging interactions using PSA and P-IoU, while Posterior Rationality assess its robustness. To ensure statistical significance, the system generates five predictions for each stressor per trial, and we report their average scores. As shown in Table \ref{tab:goal}, PSA (semantic alignment) decreases slightly compared to the main study (Table \ref{tab:quantitative}), reflecting that unconventional user interventions make intent prediction inherently more difficult. Note that for Stressor C, semantic-based metrics are excluded, as these "destructive" instructions intentionally lack a constructive subject; however, users still provide a bounding box to guide the robot's spatial response. In contrast, P-IoU remains stable and Rationality scores stay high, suggesting that CoBrush maintains coherent spatial decisions and plausible behavior even under duress. Although performance shows a degradation from Stressor A to C, the overall drop is marginal, suggesting that the proposed framework is relatively insensitive to the increased ambiguity and conflict introduced in the stress test.

\begin{table}[t]
\centering
\caption{Results of the Stress Test evaluating adaptability and robustness under increasing perturbation levels.}
\label{tab:stressors}
\begin{tabular}{lcccc}
\toprule
 & PSA  & P-IoU  & Post. target  & Post. bbox  \\
\midrule
A-Mild & 0.555 & 0.354 & 7.14 & 7.39 \\
B-Moderate & 0.538 & 0.353 & 6.81 & 6.86 \\
C-Severe & --    & 0.385 & --   & 6.77 \\
\bottomrule
\end{tabular}
\label{tab:goal}
\vspace{-1mm}
\end{table}

\subsection{Runtime Breakdown}

We report the average runtime of each pipeline component across multiple interaction rounds. Semantic planning remains lightweight, with semantic abstraction and coarse grounding requiring approximately 71–96\,s per round, Visual synthesis and Coarse-to-fine localization takes 52–88\,s in total, including about 9.5\,s for region refinement. In contrast, physical execution dominates the overall runtime. Stroke rendering and robotic arm execution require 22–63\,min per round, depending on the number of strokes and color switches. On average, the robotic arm operates at approximately 3\,min per 10 strokes. A typical co-painting session with 4–5 interaction rounds lasts 4–5\,hours.

\subsection{User Experience}
% \noindent\textbf{User experience. }
Participants rated their experience across five dimensions on a 10-point Likert scale (values reported as $Mean$, $SD$). Overall, the results indicate a favorable and low-effort collaboration, with high satisfaction ($7.11, 1.17$) and minimal perceived workload ($2.33, 1.22$). While participants found the turn-taking flow ($5.56, 2.35$) and creative support ($5.56, 2.07$) to be consistently functional, the moderate ratings for intent alignment ($6.33, 1.41$) suggest that future improvements should focus on fine-grained semantic grounding. These findings confirm CoBrush's usability while highlighting the inherent challenge of capturing subtle artistic nuances in open-ended co-painting.

\section{Conclusion and Future work}
This work takes a first step toward viewing embodied co-painting as a structured decision-making problem. We show that a hierarchical organization provides a practical and effective way to support coherent human–robot collaboration on a real physical canvas, offering  both stability and flexibility during iterative interaction. Beyond the current system, this framework naturally opens several directions for future exploration. One promising path is to introduce adaptive decision policies that can learn when and how different components of the pipeline should be invoked, enabling more strategic control over painting progress. Another is to model user-specific preferences and painting rhythms, allowing the agent to personalize its behavior to different creative styles and collaboration patterns. In addition, extending evaluation to more semantically complex themes, such as multi-object compositions and abstract concepts, can further test the generality of the proposed structure. Overall, this work establishes a concrete starting point for developing more adaptive and personalized co-painting systems, and shows that structured decision processes are a practical basis for stable and flexible human–robot collaboration on real physical canvases.



\section*{ACKNOWLEDGMENT}

This work is supported by the Research Grants Council of Hong Kong under the Theme-based Research Scheme (T45-205/21-N).

%%%%%%%%%%%%%%%%%%%%%%%%%%%%%%%%%%%%%%%%%%%%%%%%%%%%%%%%%%%%%%%%%%%%%%%%%%%%%%%%

\bibliographystyle{unsrt}
\bibliography{copaint}

@inproceedings{oh2018lead,
  title={I lead, you help but only with enough details: Understanding user experience of co-creation with artificial intelligence},
  author={Oh, Changhoon and Song, Jungwoo and Choi, Jinhan and Kim, Seonghyeon and Lee, Sungwoo and Suh, Bongwon},
  booktitle={Proceedings of the 2018 CHI conference on human factors in computing systems},
  pages={1--13},
  year={2018}
}

@inproceedings{davis2015drawing,
  title={Drawing apprentice: An enactive co-creative agent for artistic collaboration},
  author={Davis, Nicholas and Hsiao, Chih-PIn and Singh, Kunwar Yashraj and Li, Lisa and Moningi, Sanat and Magerko, Brian},
  booktitle={Proceedings of the 2015 ACM SIGCHI Conference on Creativity and Cognition},
  pages={185--186},
  year={2015}
}

@inproceedings{davis2016empirically,
  title={Empirically studying participatory sense-making in abstract drawing with a co-creative cognitive agent},
  author={Davis, Nicholas and Hsiao, Chih-PIn and Yashraj Singh, Kunwar and Li, Lisa and Magerko, Brian},
  booktitle={Proceedings of the 21st International Conference on Intelligent User Interfaces},
  pages={196--207},
  year={2016}
}

@inproceedings{karimi2019relating,
  title={Relating cognitive models of design creativity to the similarity of sketches generated by an ai partner},
  author={Karimi, Pegah and Davis, Nicholas and Maher, Mary Lou and Grace, Kazjon and Lee, Lina},
  booktitle={Proceedings of the 2019 Conference on Creativity and Cognition},
  pages={259--270},
  year={2019}
}

@inproceedings{guljajeva2022dream,
  title={Dream painter: an interactive art installation bridging audience interaction, robotics, and creative AI},
  author={Guljajeva, Varvara and Canet Sola, Mar},
  booktitle={Proceedings of the 30th ACM international conference on multimedia},
  pages={7235--7236},
  year={2022}
}

@inproceedings{schaldenbrand2024cofrida,
  title={Cofrida: Self-supervised fine-tuning for human-robot co-painting},
  author={Schaldenbrand, Peter and Parmar, Gaurav and Zhu, Jun-Yan and McCann, James and Oh, Jean},
  booktitle={2024 IEEE International Conference on Robotics and Automation (ICRA)},
  pages={2296--2302},
  year={2024},
  organization={IEEE}
}

@article{ibarrola2023collaborative,
  title={A collaborative, interactive and context-aware drawing agent for co-creative design},
  author={Ibarrola, Francisco and Lawton, Tomas and Grace, Kazjon},
  journal={IEEE Transactions on Visualization and Computer Graphics},
  volume={30},
  number={8},
  pages={5525--5537},
  year={2023},
  publisher={IEEE}
}

@article{ha2017neural,
  title={A neural representation of sketch drawings},
  author={Ha, David and Eck, Douglas},
  journal={arXiv preprint arXiv:1704.03477},
  year={2017}
}

@inproceedings{radford2021learning,
  title={Learning transferable visual models from natural language supervision},
  author={Radford, Alec and Kim, Jong Wook and Hallacy, Chris and Ramesh, Aditya and Goh, Gabriel and Agarwal, Sandhini and Sastry, Girish and Askell, Amanda and Mishkin, Pamela and Clark, Jack and others},
  booktitle={International conference on machine learning},
  pages={8748--8763},
  year={2021},
  organization={PmLR}
}

@article{frans2022clipdraw,
  title={Clipdraw: Exploring text-to-drawing synthesis through language-image encoders},
  author={Frans, Kevin and Soros, Lisa and Witkowski, Olaf},
  journal={Advances in Neural Information Processing Systems},
  volume={35},
  pages={5207--5218},
  year={2022}
}

@article{pan2025omnimanip,
  title={OmniManip: Towards General Robotic Manipulation via Object-Centric Interaction Primitives as Spatial Constraints},
  author={Pan, Mingjie and Zhang, Jiyao and Wu, Tianshu and Zhao, Yinghao and Gao, Wenlong and Dong, Hao},
  journal={arXiv preprint arXiv:2501.03841},
  year={2025}
}

@proceedings{ziliotto2025tango,
  title={TANGO: Training-free Embodied AI Agents for Open-world Tasks},
  author={Ziliotto, Filippo and Campari, Tommaso and Serafini, Luciano and Ballan, Lamberto},
  booktitle={Proceedings of the Computer Vision and Pattern Recognition Conference},
  pages={24603--24613},
  year={2025}
}

@article{feng2023layoutgpt,
  title={Layoutgpt: Compositional visual planning and generation with large language models},
  author={Feng, Weixi and Zhu, Wanrong and Fu, Tsu-jui and Jampani, Varun and Akula, Arjun and He, Xuehai and Basu, Sugato and Wang, Xin Eric and Wang, William Yang},
  journal={Advances in Neural Information Processing Systems},
  volume={36},
  pages={18225--18250},
  year={2023}
}

@inproceedings{lin2025elements,
  title={From Elements to Design: A Layered Approach for Automatic Graphic Design Composition},
  author={Lin, Jiawei and Sun, Shizhao and Huang, Danqing and Liu, Ting and Li, Ji and Bian, Jiang},
  booktitle={Proceedings of the Computer Vision and Pattern Recognition Conference},
  pages={8128--8137},
  year={2025}
}

@inproceedings{tang2025ata,
  title={ATA: Adaptive Transformation Agent for Text-Guided Subject-Position Variable Background Inpainting},
  author={Tang, Yizhe and Sun, Zhimin and Du, Yuzhen and Yi, Ran and Lu, Guangben and Hu, Teng and Li, Luying and Ma, Lizhuang and Zou, Fangyuan},
  booktitle={Proceedings of the Computer Vision and Pattern Recognition Conference},
  pages={18335--18345},
  year={2025}
}

@inproceedings{xue2025comfybench,
  title={Comfybench: Benchmarking llm-based agents in comfyui for autonomously designing collaborative ai systems},
  author={Xue, Xiangyuan and Lu, Zeyu and Huang, Di and Wang, Zidong and Ouyang, Wanli and Bai, Lei},
  booktitle={Proceedings of the Computer Vision and Pattern Recognition Conference},
  pages={24614--24624},
  year={2025}
}

@inproceedings{vinker2025sketchagent,
  title={Sketchagent: Language-driven sequential sketch generation},
  author={Vinker, Yael and Shaham, Tamar Rott and Zheng, Kristine and Zhao, Alex and E Fan, Judith and Torralba, Antonio},
  booktitle={Proceedings of the Computer Vision and Pattern Recognition Conference},
  pages={23355--23368},
  year={2025}
}

@inproceedings{gao2020making,
  title={Making robots draw a vivid portrait in two minutes},
  author={Gao, Fei and Zhu, Jingjie and Yu, Zeyuan and Li, Peng and Wang, Tao},
  booktitle={2020 IEEE/RSJ International Conference on Intelligent Robots and Systems (IROS)},
  pages={9585--9591},
  year={2020},
  organization={IEEE}
}

@inproceedings{luo2018robot,
  title={Robot artist performs cartoon style facial portrait painting},
  author={Luo, Ren C and Liu, Yu Jung},
  booktitle={2018 IEEE/RSJ International Conference on Intelligent Robots and Systems (IROS)},
  pages={7683--7688},
  year={2018},
  organization={IEEE}
}

@article{low2022drozbot,
  title={drozbot: Using ergodic control to draw portraits},
  author={L{\"o}w, Tobias and Maceiras, J{\'e}r{\'e}my and Calinon, Sylvain},
  journal={IEEE Robotics and Automation Letters},
  volume={7},
  number={4},
  pages={11728--11734},
  year={2022},
  publisher={IEEE}
}

@inproceedings{gao2024human,
  title={Human-Robot Interactive Creation of Artistic Portrait Drawings},
  author={Gao, Fei and Dai, Lingna and Zhu, Jingjie and Du, Mei and Zhang, Yiyuan and Qiao, Maoying and Xia, Chenghao and Wang, Nannan and Li, Peng},
  booktitle={2024 IEEE International Conference on Robotics and Automation (ICRA)},
  pages={11297--11304},
  year={2024},
  organization={IEEE}
}

@article{jin2023semi,
  title={A Semi-automatic Oriental Ink Painting Framework for Robotic Drawing from 3D Models},
  author={Jin, Hao and Lian, Minghui and Qiu, Shicheng and Han, Xuxu and Zhao, Xizhi and Yang, Long and Zhang, Zhiyi and Xie, Haoran and Konno, Kouichi and Hu, Shaojun},
  journal={IEEE Robotics and Automation Letters},
  volume={8},
  number={10},
  pages={6667--6674},
  year={2023},
  publisher={IEEE}
}

@article{liu2021robust,
  title={Robust robotic 3-D drawing using closed-loop planning and online picked pens},
  author={Liu, Ruishuang and Wan, Weiwei and Koyama, Keisuke and Harada, Kensuke},
  journal={IEEE Transactions on Robotics},
  volume={38},
  number={3},
  pages={1773--1792},
  year={2021},
  publisher={IEEE}
}

@article{song2023ssk,
  title={SSK: Robotic pen-art system for large, nonplanar canvas},
  author={Song, Daeun and Park, Jiyoon and Kim, Young J},
  journal={IEEE Transactions on Robotics},
  volume={39},
  number={4},
  pages={3106--3119},
  year={2023},
  publisher={IEEE}
}

@article{zhang2019intelligent,
  title={Intelligent Chinese calligraphy beautification from handwritten characters for robotic writing},
  author={Zhang, Xinyue and Li, Yuanhao and Zhang, Zhiyi and Konno, Kouichi and Hu, Shaojun},
  journal={The Visual Computer},
  volume={35},
  pages={1193--1205},
  year={2019},
  publisher={Springer}
}

@inproceedings{lindemeier2015hardware,
  title={Hardware-Based Non-Photorealistic Rendering Using a Painting Robot},
  author={Lindemeier, Thomas and Metzner, Jens and Pollak, Lena and Deussen, Oliver},
  booktitle={Computer graphics forum},
  volume={34},
  number={2},
  pages={311--323},
  year={2015},
  organization={Wiley Online Library}
}

@article{scalera2019watercolour,
  title={Watercolour robotic painting: a novel automatic system for artistic rendering},
  author={Scalera, Lorenzo and Seriani, Stefano and Gasparetto, Alessandro and Gallina, Paolo},
  journal={Journal of Intelligent \& Robotic Systems},
  volume={95},
  pages={871--886},
  year={2019},
  publisher={Springer}
}

@inproceedings{schaldenbrand2023frida,
  title={Frida: A collaborative robot painter with a differentiable, real2sim2real planning environment},
  author={Schaldenbrand, Peter and McCann, James and Oh, Jean},
  booktitle={2023 IEEE International Conference on Robotics and Automation (ICRA)},
  pages={11712--11718},
  year={2023},
  organization={IEEE}
}

@article{oquab2023dinov2,
  title={Dinov2: Learning robust visual features without supervision},
  author={Oquab, Maxime and Darcet, Timoth{\'e}e and Moutakanni, Th{\'e}o and Vo, Huy and Szafraniec, Marc and Khalidov, Vasil and Fernandez, Pierre and Haziza, Daniel and Massa, Francisco and El-Nouby, Alaaeldin and others},
  journal={arXiv preprint arXiv:2304.07193},
  year={2023}
}

@article{liu2023visual,
  title={Visual instruction tuning},
  author={Liu, Haotian and Li, Chunyuan and Wu, Qingyang and Lee, Yong Jae},
  journal={Advances in neural information processing systems},
  volume={36},
  pages={34892--34916},
  year={2023}
}

@article{ren2024grounded,
  title={Grounded sam: Assembling open-world models for diverse visual tasks},
  author={Ren, Tianhe and Liu, Shilong and Zeng, Ailing and Lin, Jing and Li, Kunchang and Cao, He and Chen, Jiayu and Huang, Xinyu and Chen, Yukang and Yan, Feng and others},
  journal={arXiv preprint arXiv:2401.14159},
  year={2024}
}

@article{huang2025t2i,
  title={T2i-compbench++: An enhanced and comprehensive benchmark for compositional text-to-image generation},
  author={Huang, Kaiyi and Duan, Chengqi and Sun, Kaiyue and Xie, Enze and Li, Zhenguo and Liu, Xihui},
  journal={IEEE Transactions on Pattern Analysis and Machine Intelligence},
  year={2025},
  publisher={IEEE}
}

@article{qin2026data,
  title={Data-Efficient Brushstroke Generation with Diffusion Models for Oil Painting},
  author={Qin, Dantong and Bozzon, Alessandro and Yang, Xian and Zhang, Xun and Guo, Yike and Wang, Pan},
  journal={ACM Transactions on Multimedia Computing, Communications and Applications},
  year={2026},
  publisher={ACM New York, NY}
}

\end{document}